\documentclass[letterpaper, 10 pt, conference]{ieeeconf}

\IEEEoverridecommandlockouts
\usepackage{graphics} 
\usepackage{epsfig} 
\usepackage{mathptmx} 
\usepackage{times} 
\usepackage{amsmath} 
\usepackage{amssymb}  
\usepackage{cite}
\usepackage{cleveref}
\usepackage{siunitx}
\usepackage{subcaption}

\usepackage{tabularx}
\Crefname{figure}{Fig.}{Figs.}
\Crefname{section}{Sec.}{Secs.}
\Crefname{table}{Tab.}{Tabs.}
\Crefname{equation}{Eqn.}{Eqns.}

\newcommand{\degree}{$^\circ$}

\newcommand{\ahip}{A_\mathrm{hip}}
\newcommand{\aknee}{A_\mathrm{knee}}
\newcommand{\ohip}{O_\mathrm{hip}}
\newcommand{\oknee}{O_\mathrm{knee}}

\newcommand{\fnet}{F_\mathrm{net}}
\newcommand{\winput}{W_\mathrm{input}}

\newcommand{\fthrust}{F_\mathrm{thrust}}
\newcommand{\df}{dorsiflexion}
\newcommand{\DuckLeg}{Duck Leg} 

\newcommand\alex  [1]{{\color{black} #1}}

\graphicspath{{Pic}}

\title{\LARGE \bf
Bird-inspired tendon coupling improves paddling \\ efficiency by shortening phase transition times}

\author{Jianfeng Lin, Zhao Guo, Alexander Badri-Spröwitz
\thanks{%
J. Lin is with the School of Physics, Georgia Institute of Technology, Atlanta, GA. 
J. Lin and Z. Guo are with the School of Power and Mechanical
Engineering, Wuhan University, Wuhan, China.
ABS is with the MPI for Intelligent Systems, Stuttgart, Germany and with the Department of Mechanical Engineering, KU Leuven, Leuven, Belgium.
Corresponding e-mail: {\tt\small jianf.lin@gatech.edu}
}}

\begin{document}
\maketitle

\begin{abstract}
Drag-based swimming with rowing appendages, fins, and webbed feet is a widely adapted locomotion form in aquatic animals. To develop effective underwater and swimming vehicles, a wide range of bioinspired drag-based paddles have been proposed, often faced with a trade-off between propulsive efficiency and versatility. Webbed feet provide an effective propulsive force in the power phase, are light weight and robust, and can even be partially folded away in the recovery phase. However, during the transition between recovery and power phase, much time is lost folding and unfolding, leading to drag and reducing efficiency. In this work, we took inspiration from the coupling tendons of aquatic birds and utilized tendon coupling mechanisms to shorten the transition time between recovery and power phase. Results from our hardware experiments show that the proposed mechanisms improve propulsive efficiency by 2.0 and 2.4 times compared to a design without extensor tendons or based on passive paddle, respectively. We further report that distal leg joint clutching, which has been shown to improve efficiency in terrestrial walking, did not play an major role in swimming locomotion. In sum, we describe a new principle for an efficient, drag-based leg and paddle design, with potential relevance for the swimming mechanics in aquatic birds.
\end{abstract}

\section{Introduction}
\begin{figure}[thpb]
\centering
\includegraphics[scale=0.7]{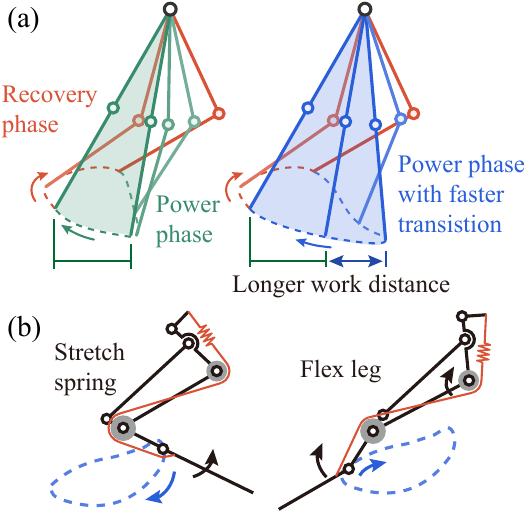}
\caption{Overview of the two hypotheses tested. (a) The torque applied through a coupling extensor tendon helps resetting the foot orientation after the recovery stroke. This decreases the phase transition time. In comparison: long transition time in green leaves little time for the power phase. A short transition time increases the time the paddle can spend in the power phase (blue). (b) We hypothesize that the global spring tendon stores energy throughout the power phase and supports flexing the leg at the end of the phase through its coupling, leading to an improved overall power phase.}
\label{fig:shortenT}
\end{figure}

Drag-based propulsion is a common strategy for  animals\cite{vogel2008modes,lock2013multi} such as krills\cite{alben2010coordination}, fishes\cite{sfakiotakis1999review}, turtles\cite{wyneken2013biology}, and aquatic birds. Specifically, the gait cycle of paddling consists of two phases; a power and a recovery phase\cite{provini2012walking}. Thrust is generated during the power phase by pushing backwards. In the recovery phase the animal pulls the foot forward in preparation for the next propulsion phase and with as low drag as possible. Between the power and the recovery phase, a short transition occurs where the foot is reoriented. Each cycle's effective thrust is the net force of both main and transition phases.
In nature, neuromechanical and morphological adaptations contribute to the optimization of aquatic locomotion capabilities\cite{crespi2013salamandra,kwak2019comprehensive,baines2022multi}. Studying the principles behind these adaptations can lead to improved engineering mechanisms with better adaptability, agility, and versatility\cite{palmisano2013power,diaz2021minimal,kwak2019comprehensive,baines2022multi,simha2020flapped,wang2024aquamilr,wang2023mechanical,chong2023multilegged}.
\begin{figure}[t]
\centering
\includegraphics[scale=.8]{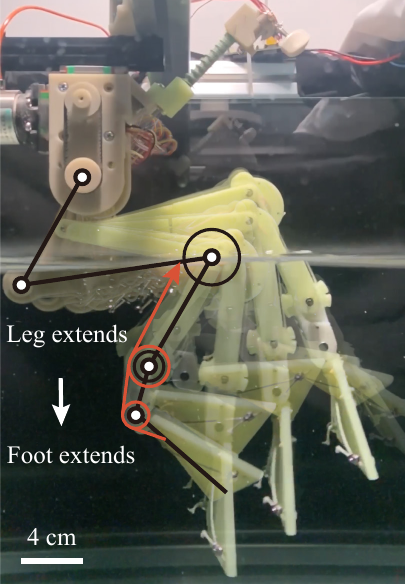}
\caption{\textbf{Picture of the robotic \DuckLeg.} Overlaid is an extensor tendon acting on the foot once the leg extends. Two brushless motors drive the leg through belts, in leg angle and leg length direction. The foot is self-folding during the recovery phase, and the extensor tendon supports its rapid extension into the power phase.}

\label{fig:prototype}
\end{figure}
State-of-art paddle designs feature either actively controlled or passive paddles. Active actuation\cite{huang2021cormorant,liu2018platform} comes with higher control complexity, has limits due to the underwater environment, and typically higher end effector inertia. Passive paddles can be designed from two linkages and a limiter of the flexible joint (\Cref{fig:shortenT}b). The elastic joint bends passively during the recovery phase and reduces drag resistance. The joint limit maximizes thrust by keeping the paddle at the position of the maximum force arm during the power phase (\Cref{fig:shortenT}a). The concept has been implemented previously in various forms and materials\cite{pham2020dynamic,simha2020flapped,kwak2019comprehensive}. As we will show with one comparison setup, the fixed joint stiffness leads to longer transition times between main phases and therefor reduces the time of the power phase (\Cref{fig:shortenT}a). Kwak et al. added a relaxation phase to prepare the paddle for the power phase \cite{kwak2019comprehensive}. Sharifzadeh et al.~optimized the anisotropic stiffness for the specific fluid environment\cite{sharifzadeh2021reconfigurable,sharifzadeh2020curvature}. However, both methods sacrifice power phase time and versatility.

We took inspiration from aquatic birds which are excellent surface water swimmer with webbed feet for effective paddling\cite{johansson2003delta}. Their morphology features a shorter tarsometatarsus and femur for efficient locomotion with thrust produced closer to the body\cite{fish1996transition,clifton2018comparative}. Birds swim efficiently with a large range of ankle and a smaller range of hip and knee motion leading to a long working distance\cite{provini2012walking}. How exactly the muscular system interacts during locomotion is less studied, also due to difficulties of in vivo experiments. For terrestrial walking locomotion, bird's extensive muscle-tendon coupling over multiple joints into feet and toes has been found to be highly efficient \cite{badri2022birdbot,chatterjee2022multi,chang2017mechanical}. Both similarities and differences exist between muscle function for terrestrial and aquatic locomotion. Gastrocnemius and iliotibialis Lateralis pars post acetabularis (ILPO) muscles feature similar functions for terrestrial and swimming locomotion\cite{biewener2001dynamics,carr2008muscle}. Here, we hypothesize that the coupling of interconnecting tendons plays a significant role in effective paddling. First, \textbf{the coupled extensor tendon helps reducing the transition between recovery and power phase by inducing a foot joint extending torque at the onset of the power phase} (\Cref{fig:shortenT,fig:prototype}). Second, leg joint clutching, which is hypothesized in terrestrial locomotion\cite{badri2022birdbot}, might also be relevant in swimming. We therefore hypothesize that the leg clutch increases efficiency by storing energy at the beginning of the power phase and flexing the leg at the end of the power phase (\Cref{fig:shortenT}b).
To verify our hypothesis, we designed the \DuckLeg~(\Cref{fig:prototype}) with a webbed foot and a network of coupling tendons. The webbed foot opens and close without actuators. The tendon network includes a global spring tendon (GST) and several spring extensor tendons. We tested different tendon configurations for the above two hypotheses and measure propulsive force, paddling efficiency, and transition times.
\begin{figure*}[t]
\centering
\includegraphics[scale=.9]{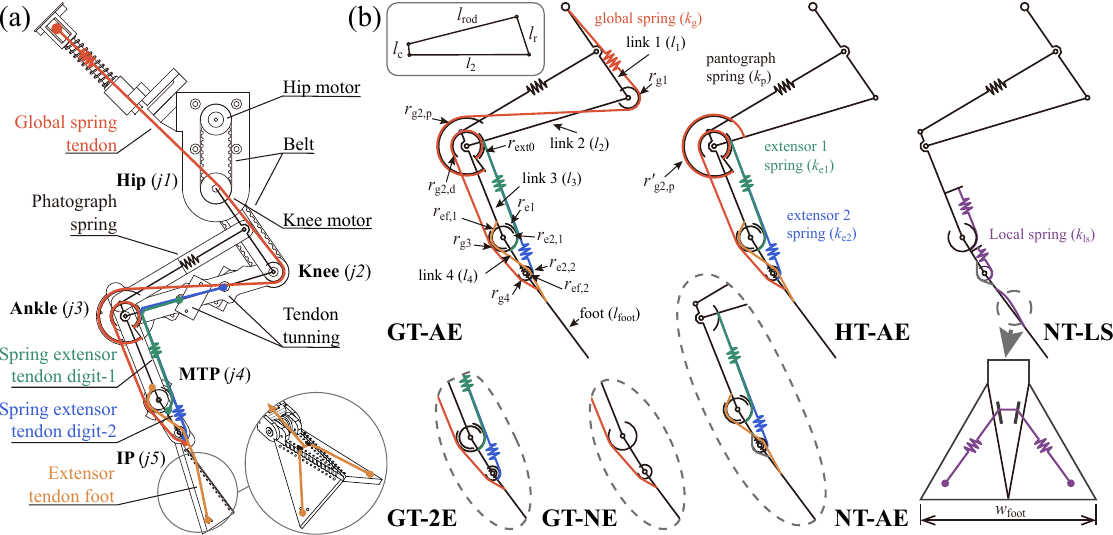}
\caption{\textbf{Design and tendon configurations of \DuckLeg.} (a) \DuckLeg's design details. Hip and knee joints are actuated by brushless motors, connecting to the joints by belts. The global spring tendon and several extensor tendons mechanically couple the remaining joints. 
(b) Schematics of tested tendon configurations, in sum three sets: the full global spring tendon (global tendon, GT), the half global spring tendon (half tendon, HT), and a set without a global spring tendon (no tendon, NT). Extensor-tendon configurations include either all extensors (AE), two extensors (2E), no extensors (NE), or a local spring (LS).}
\label{fig:tendonconfig}
\end{figure*}

This work contributes two novel findings: through deploying different tendon arrangements, we identified one function of coupling tendons in swimming locomotion and how they improve effectiveness when compared to traditional passive paddles. By measuring the transition time, we confirmed that our proposed mechanism increases  paddling efficiency. Our work also provides a potential new understanding of the biomechanics of aquatic birds. 

\section{Methods}
\textbf{DuckLeg Design:} We designed \DuckLeg~about duck sized (\Cref{tab:designpara}), and with the option to implement different tendon configurations. The main leg structure includes five joints (\alex{\Cref{fig:tendonconfig}}). The hip and knee of \DuckLeg~are actuated by timing belts, similar to the SOLO robot\cite{grimminger2020open}. The ankle was coupled to the knee through an irregular four-bar link, the transmission ratio between the ankle and the knee is approximately 4 to match the large ankle amplitude that birds exhibit during swimming\cite{messinger1979mechanics,biewener2001dynamics,carr2008muscle}. The two most distal  segments include a link with \SI{10}{\degree} extension limitation and a passive webbed foot. The webbed foot is an equilateral triangle divided into three parts. The dividing line passes through the midpoint of the bottom edge. The different triangular segments are connected at the bottom by nylon cloth and Kevlar rope inspired by previous designs\cite{hepp2022novel,lin2024bioinspired}, enabling free rotation and featuring a hard stop ('joint limit') in the opposite direction. Link lengths and foot parameters (\Cref{tab:designpara}) are based on the morphological sizes of semi-aquatic birds \cite{clifton2018comparative,taylor2020waddle,johansson2003delta,kim2011characteristics}.

MTP, IP, and foot joints are coupled to the leg through the global spring tendon (GST) and extensor tendons (or cams~\cite{badri2022birdbot}). Tendon pulleys (\Cref{fig:tendonconfig} $\df$) are simplifications of joints' sesamoid bones, specifically patella~\cite{regnault2017}. Pulley radii determine transmit tendon force into joint torque \cite{prilutsky1994tendon}. We used the same design principles as BirdBot\cite{badri2022birdbot}, determining the pulley radii via equal joint effective mechanical advantages (EMA)\cite{biewener1989scaling} in standing position to create a leg that could potentially allow standing. The coupling kinematics of the extensor tendon in the power phase can be treated as a drive from the ankle to a digit joint with the transmission ratio $n_j$:
\begin{equation}
\Delta \theta_{j}=n_j*\Delta \theta_{\rm{ankle}}, \;\;\; j=\{\rm{MTP}, \rm{IP}\},
\end{equation}
where the transmission ratio is determined by the radii of pulleys that the tendon crosses. Here, while ensuring that no antagonism occurs between the extensor tendons and GST, the transmission ratio $n_j$ is designed as high as possible (i.e., $n_{\rm{MTP}}=1.6$). 
\begin{figure}[thpb]
\centering
\includegraphics[scale=1]{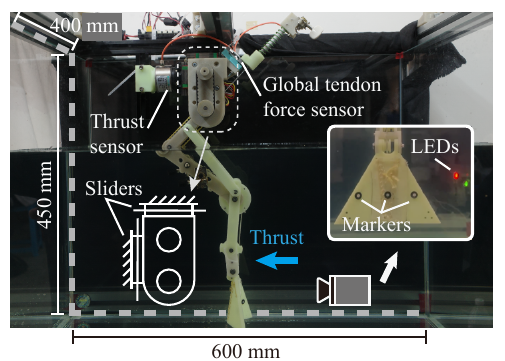}
\caption{\textbf{The experimental platform of \DuckLeg.}}
\label{fig:expset}
\end{figure}

We designed six different tendon configurations by adjusting tendon attachments of \DuckLeg~(\Cref{fig:tendonconfig}). To verify a potential shortened transition from recovery to power phase, we modulated extensor tendons (GT: global tendon):
\begin{itemize}
\item \textbf{GT-AE}: All extensor tendons (abbreviated AE) are connected to the leg.
\item \textbf{GT-2E}: The spring extensor tendon digit 1 and 2 are connected to the leg (2E, two extensor tendons are connected). 
\item \textbf{GT-NE}: No extensor tendons (abbreviated NE) are connected to the leg.
\end{itemize}
The performance between GT-2E and GT-NE could demonstrate the function of extensor tendons. Second, to discover the function of GST on aquatic locomotion, we kept the all-extensor-connected configuration (abbreviated AE) and designed the other two configurations:
\begin{itemize}
\item \textbf{HT-AE}: Only half the GST has remained (proximal part, abbreviated AE).
\item \textbf{NT-AE}: No global tendon is connected (abbreviated NT). Instead, the IP joint is limited by a rigid rope.
\end{itemize}
GT-AE couples the foot joints to the knee, which distributes the torque among all the joints of the leg. HT-AE only connects to the ankle, which has an in-series connection with the knee. By comparing these two configurations, we can discover the role of such global parallel elasticity during swimming. Comparison between HT-AE and NT-AE can provide insight into the necessity of toe coupling for swimming. To ensure that the coupling kinematics is the same between HT-AE and GT-AE (half of GST is removed), we recalculated the radius of the GST pulley as $r_{g2,p}^{'}$. We also designed a configuration according to the conventional design of a passive paddle without GST and local spring joints (NT-LS). Design parameters can be found in \Cref{tab:designpara}.

\begin{figure*}[thpb]
\centering
\includegraphics[scale=1]{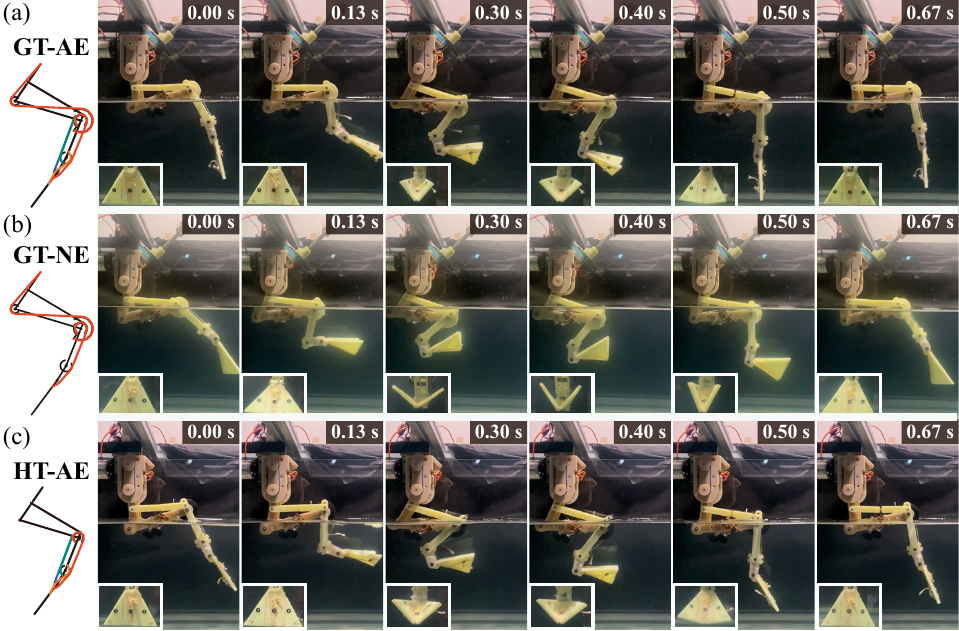}
\caption{\textbf{Snapshots of back and side view of paddling gaits of GT-AE, GT-NE, and HT-AE.} Compared to GT-NE, which does not have connected extensor tendons, GT-AE opens the foot faster (at \SI{0.50}{s}, the foot is almost fully open). HT-AE has the similar foot opening time due to the similar extensor configurations, but the ankle displacement is relatively small compared to GT-AE.}

\label{fig:gaits}
\end{figure*}
\textbf{Experimental Setup:}
The \DuckLeg~is fixed on a 600 by 400 by \SI{450}{mm} sized water tank with an aluminum frame (\Cref{fig:expset}). Water was added until it covered the knee joint. Two sliders are connected at the top and front of the \DuckLeg~to decouple horizontal and vertical forces. A thrust sensor (one-axis load cell, HYMN-019, 100N) is fixed to the front slider. A miniature load cell (FMZK-1KG, 100N) is connected between the body and the global spring to record the global tendon force. A high-speed camera (240 frames per second) records the side view of the webbed foot to allow analyzing the foot opening. Three makers are attached on each part of the webbed foot, with positions in line when the foot fully opens.  In the side view, red and green LEDs are placed to indicate the gait phase and the program status.
The two brushless motors are PID position controlled, programmed on a microcontroller (RoboMaster Development Board Type A). A custom-written LabView program on a compactRIO controller (sbRIO-9637, National Instruments) synchronizes gait and sensor information. The data of two force sensors is sampled by the system at $f=\SI{800}{Hz}$, with analog transmitters (S641C-P, $0\pm 10$V output). Further, current and motor angles of two motors are recorded. The system generates gait signals sent to the low-level microcontroller and synchronizes the LEDs with the gait phase. We implemented a central pattern generator (CPG) to generate hip and knee trajectories. CPGs are a common controller to replicate the rhythmic motion of vertebrates\cite{ijspeert2007swimming}:
\begin{equation}
\begin{aligned}
&\theta_{\rm{Hip}}=A_{\rm{Hip}}\rm{cos}(\Theta)+O_{\rm{Hip}}  \\
&\theta_{\rm{Knee}}=A_{\rm{Knee}}\rm{cos}(\Theta+\varphi)+O_{\rm{Knee}} \\
\end{aligned},
\end{equation}
\begin{equation}
\Theta = \left\{\begin{matrix}
\begin{aligned}
&\Phi /2D_{vir} &\Phi<2\pi D_{vir}\\
&(\Phi +2\pi (1-2D_{vir}))/2(1-D_{vir}) &\mathrm{else}\\
\end{aligned},
\end{matrix}\right.
\end{equation}
where $\ahip$ and $\aknee$ represent the amplitude of the hip and knee, $\ohip$ and $\oknee$ are the offset of the hip and knee, $\varphi$ is the phase difference between two joints, and $\Phi$ represents the oscillator's linearly progressing phase. The gait is executed at \SI{1.5}{Hz}.
   
\textbf{Net Thrust and Propulsive Efficiency:} To compare the different configurations of the tendon, we chose the net thrust and propulsive efficiency to evaluate paddling performance. Net force refers to the difference between resistance force and propulsive force in one gait cycle, based on which we evaluate the external work\cite{sharif2021}:
\begin{equation}
\fnet = \frac{1}{T}\int_{0}^{T}{ \fthrust }(t)dt 
\end{equation}
where $T$ denotes to the period of one gait cycle, $\fthrust$ is the instantaneous thrust of the paddle which can be recorded by load cell.
Propulsive efficiency $\eta=W_{\rm{use}}/W_{\rm{input}}$ is the portion of the useful propulsive power $W_{\rm{use}}$ to the total input energy $W_{\rm{input}}$. For the paddle fixed into the water\cite{palmisano2013maximize}, the propulsive efficiency can be simplified to:
\begin{equation}
\eta=\frac{\fnet}{\winput} = \frac{\fnet}{ \int_{0}^{T} VI(t)dt}
\end{equation}
where $V$ and $I(t)$ are voltage and current, respectively.

\begin{figure*}[t]
\centering
\includegraphics[scale=.9]{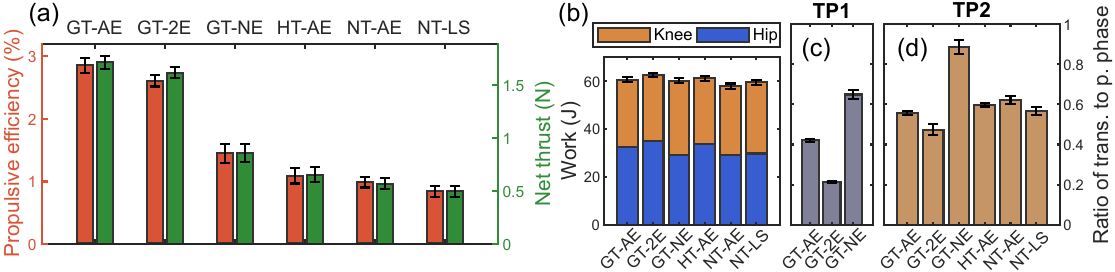}
\caption{\textbf{Summary of experiment results.} Propulsive efficiency and net thrust (a), and motor input work (b) during the gait cycle for different tendon configurations. The ratios of transition time of TP1 (c) and TP2 (d) to power phase timing are listed.}

\label{fig:summary}
\end{figure*}

\textbf{Phase Transition Time Analysis:}
To analyze the transition time from the recovery phase to the power phase, we choose two quantities. First, the time when the force on GST first appeared was chosen and named Transition Point 1 (TP1). When the foot begins to perform work, the GST stops the foot from overextending. Time comparison can identify whether the phase transition is shortened.  This measure is only suitable for tendon configurations that feature a global tendon, such as GT-AE, GT-2E, and GT-NE.
Second, we measure the time when the foot is fully open as another value to analyze the transition time (Transition Point 2, TP1). The open position of the webbed foot helps quantifying transition time; a foot that does not fully open until late in the gait cycle has a longer transition time. Three markers were placed on the foot (\Cref{fig:expset}), and when fully open, the markers align in a straight line. This method works without strict camera positioning, except when the foot is vertical to the camera, which can be filtered by a threshold.

\section{Results}
\textbf{Summary}: The summary of results of propulsive efficiency, net thrust, motor work, and transition time is shown in \Cref{fig:summary}. The error bar represents the 95$\%$ confidence interval from 100 averaged gait cycles. \Cref{fig:gaits} shows snapshots of the paddling gait of GT-AE, GT-NE, and HT-AE. The gait of other tendon configurations can be found in the supplementary video.
The NT-LS group, based on the traditional passive paddle design, has has the lowest propulsive efficiency ($0.84\pm0.09\%$) within the six tendon configurations.
The configuration with intact tendons (GT-AE) demonstrates the highest propulsive efficiency ($2.86\pm0.12\%$), which is 2.39 times higher than NT-LS. With the addition of extensor tendons, propulsive efficiency improved in general. These improvements provide a first evidence for the importance of coupling tendons in improving swimming efficiency.
\Cref{fig:summary} shows that the propulsive efficiency and net thrust is increasing with the addition of coupling tendons. Motor work (\Cref{fig:summary}b, \alex{'input work'}) within different tendon configurations does not fluctuate much. These two results indicate that the efficiency improvement is likely due to a kinematic difference rather than motor energy efficiency. This implies that our second hypothesis, leg clutching, likely does contribute to efficient paddling. However, the complete GST group (GT-AE) still shows higher efficiency than the groups without GST (HT-AE and NT-AE). This may be due to the role of GST in gait adaptation, which will be further discussed in the following sections.
\begin{figure*}[thpb]
\centering
\includegraphics[scale=.95]{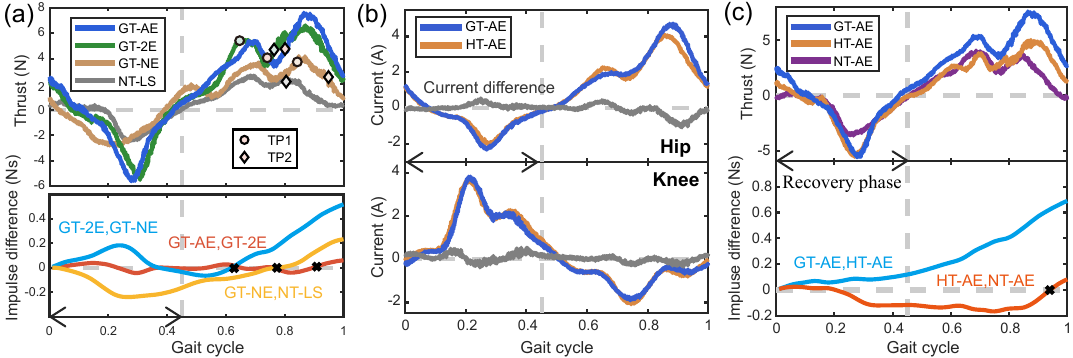} 
\caption{\textbf{Detailed analysis of shortened transition effect and GST function.}
(a) The thrust and impulse difference for GT-AE, GT-2E, GT-NE, and NT-LS. (b) The current variation for hip and knee motors during one gait cycle. (c) The thrust and impulse difference for GT-AE, HT-AE, and NT-AE.
}
\label{fig:analysis}
\end{figure*}

As expected, with the addition of extensor tendons, propulsive efficiency improved by 97$\%$ and 99$\%$ between GT-AE ($2.86\pm0.12\%$ and $1.72\pm0.06$ N) and GT-NE ($1.45\pm0.15\%$ and $0.86\pm0.09$N). The transition phase changed considerably. Compared to GT-NE, the transition times of GT-2E reduced much: TP1 was reduced by  67$\%$ and TP2 by 45$\%$. \Cref{fig:gaits} shows an intuitive comparison. For GT-AE, the foot is almost fully opened at \SI{0.50}{s}, but the GT-NE foot is still closed (\Cref{fig:gaits}b). We observed an exception of the extensor tendon function between GT-AE and GT-2E. When adding the digit tendon, the transition time increased, but an improvement in efficiency is observed. This indicates a mechanisms for efficient paddling which will be further analyzed in the following sections.

\textbf{Shortened Transition}:
As discussed in the previous part, the addition of extensor tendons leads to an improvement in paddling efficiency. To further analyze the trade-off, we plotted the thrust curve within one gait cycle (\Cref{fig:analysis}a). Data is averaged by a moving window filter for 100 gait cycles (20 gait cycles per trial repeated 5 times) for each tendon configuration. The TP1 and TP2 of the corresponding configuration are marked on the curve. On the basis of the thrust curve, the impulse difference between two tendon configurations is calculated. 
Configurations with extensor tendons have a higher negative peak in the recovery phase (\Cref{fig:analysis}a). This indicates a sacrifice when adding the extensor tendons, which results in increased resistance in the recovery phase. This sacrifice is minor; at the end of the recovery phase, the impulse differences between GT-2E and GT-NE, and GT-AE and GT-2E are nearly zero (\Cref{fig:analysis}a). The shortened transition compensates for the trade-off. The Impulse Difference between GT-2E and GT-NE becomes positive and continues to increase in the early power phase.

We found that GT-AE has a higher propulsive efficiency but a longer transition time compared to GT-2E. The Impulse Difference curve (\Cref{fig:analysis}a) shows that for most gait cycles both feature a similar thrust output. The impulse of GT-AE becomes higher than that of GT-2E near the end of the power phase, which is close to the second thrust peak of the power phase. GT-AE has the shortest time between TP1 and TP2, which could contribute to the efficiency improvement. The digit tendon couples the webbed foot with the leg. GT-2E which has a faster TP1 and a similar TP2 with GT-AE, stores the energy between TP1 and TP2 to GST and releases it at the end of the power phase when the foot has a slow rotational speed, which is wasting the stored energy. However, for GT-AE, after the GST is charged and ready to produce thrust (TP1), part of the energy is used to open the foot. The digit tendon coupling enables the leg to better use the stored energy.

\textbf{GST Function}: We had hypothesized that GST would have a similar function as in BirdBot\cite{badri2022birdbot}: storing energy at the beginning of the power phase and flexing the leg at the end of the power phase, benefiting energy efficiency. However, our results do not show energetic benefits for hip and knee motors ('input work', \Cref{fig:summary}b). 
\Cref{fig:analysis}b ('current difference') shows the current in relation to the gait cycle and the current difference between HT-AE and GT-AE, without a visible distinct difference.

We conclude that the role of the global tendon appears to be gait adaptation rather than clutching. In the thrust curve (\Cref{fig:analysis}c), three tendon configurations show the same trend, but GT-AE has the highest thrust value. In addition, the impulse difference between GT-AE and HT-AE is a proportional increase, which means that GT-AE has a better gait throughout the cycle.
An adaptive gait cycle is achieved by improving the stiffness of the ankle joint by coupling the ankle to the knee. In HT-AE and NT-AE, ankle torque during the power phase is balanced only by the pantograph spring. In GT-AE, both the global tendon (linking the knee to the IP) and the pantograph spring produce this torque. Low ankle stiffness increases ankle flexion, which leads to reduced work distance during the power phase. This indicates that in paddling locomotion, higher compliance is not as beneficial as in terrestrial locomotion\cite{kim2011characteristics}.
Despite the coupled kinematics, the global tendon also reduces the effect on individual joints by distributing torque like a multi-joint muscle\cite{prilutsky1994tendon}. Compared to HT-AE, the thrust of NT-AE reduces rapidly at the end of the power phase. When the joints are loaded, only a small portion of the energy is transmitted to the pantograph spring, as rapid dissipation of energy occurs at hard-stop joints (MTP and IP). In summary, the GST plays a role in generating a better paddling gait by mechanically coupling multiple joints, enhancing joint stiffness, and distributing the torque across joints. We found not indication that leg clutching was beneficial, unlike in terrestrial locomotion.

\section{Conclusion} 
We designed a robotic leg to investigate the propulsion efficiency of a webbed foot during the power and recovery phase of aquatic paddling, and especially during the transition between both phases. Our experiments and design variations of tendon configuration were build around two hypotheses. First, we verified that coupling of joints through extensor tendons did shorten the phase transition between the recovery and the power phase, leading to a more efficient paddling gait with reduced phase transition. The configuration with all extensor tendons (GT-AE) was 2.0 and 2.4 times more efficient, when compared to configurations without extensor tendons or a traditional passive paddle design. Second, we found that leg joint clutching powered by the global spring tendon (GST) was not beneficial during paddling, which is not consistent with our hypothesis. However, the full global spring tendon configuration demonstrated an efficiency improvement compared to the configuration with a partial or no global spring tendon. Leg and toe coupled by the GST led to a better paddling gait by improving ankle joint stiffness and distributing torque among joints. In sum, we showcased a novel and efficient leg and webbed foot for drag-based underwater vehicles. Our result also hint at the advantages of such mechanisms in their model animals: paddling semi- and aquatic birds.

\begin{table}
\centering
\begin{tabular}{c|cc|c}
Parameter&Value (mm)&Parameter&Value (mm)  \\\hline
$l_{\rm{1}}$  &79 &  $r_{\rm{g1}}$ & 7.5\\
$l_{\rm{2}}$  &120& $r_{\rm{g2,p}}$ &20 \\
$l_{\rm{3}}$  &70 & $r_{\rm{g2,d}}$ &10  \\
$l_{\rm{4}}$  &30 & $r_{\rm{g3}}$ &10 \\
$l_{\rm{foot}}$ &75 & $r_{\rm{g4}}$ &5     \\
$l_{\rm{rod}}$ &112  & $r_{\rm{ext0}}$ &12.5      \\
$l_{\rm{c}}$  &11 & $r_{\rm{e2,1}}$ &8       \\
$l_{\rm{r}}$ &40 & $r_{\rm{e2,2}}$ &3      \\

\end{tabular}
\caption{\label{tab:designpara}The parameters of \DuckLeg, defined in \Cref{fig:tendonconfig}.}
\end{table}

\bibliographystyle{ieeetr}
\bibliography{Reference}

\begin{thebibliography}{10}

\bibitem{vogel2008modes}
S.~Vogel, ``Modes and scaling in aquatic locomotion,'' {\em Integrative and
  Comparative Biology}, vol.~48, no.~6, pp.~702--712, 2008.

\bibitem{lock2013multi}
R.~Lock, S.~Burgess, and R.~Vaidyanathan, ``Multi-modal locomotion: from animal
  to application,'' {\em Bioinspiration \& biomimetics}, vol.~9, no.~1,
  p.~011001, 2013.

\bibitem{alben2010coordination}
S.~Alben, K.~Spears, S.~Garth, D.~Murphy, and J.~Yen, ``Coordination of
  multiple appendages in drag-based swimming,'' {\em Journal of The Royal
  Society Interface}, vol.~7, no.~52, pp.~1545--1557, 2010.

\bibitem{sfakiotakis1999review}
M.~Sfakiotakis, D.~M. Lane, and J.~B.~C. Davies, ``Review of fish swimming
  modes for aquatic locomotion,'' {\em IEEE Journal of oceanic engineering},
  vol.~24, no.~2, pp.~237--252, 1999.

\bibitem{wyneken2013biology}
J.~Wyneken, K.~J. Lohmann, and J.~A. Musick, {\em The biology of sea turtles},
  vol.~3.
\newblock CRC press, 2013.

\bibitem{provini2012walking}
P.~Provini, P.~Goupil, V.~Hugel, and A.~Abourachid, ``Walking, paddling,
  waddling: 3 d kinematics anatidae locomotion (c allonetta leucophrys),'' {\em
  Journal of Experimental Zoology Part A: Ecological Genetics and Physiology},
  vol.~317, no.~5, pp.~275--282, 2012.

\bibitem{crespi2013salamandra}
A.~Crespi, K.~Karakasiliotis, A.~Guignard, and A.~J. Ijspeert, ``Salamandra
  robotica ii: an amphibious robot to study salamander-like swimming and
  walking gaits,'' {\em IEEE Transactions on Robotics}, vol.~29, no.~2,
  pp.~308--320, 2013.

\bibitem{kwak2019comprehensive}
B.~Kwak, D.~Lee, and J.~Bae, ``Comprehensive analysis of efficient swimming
  using articulated legs fringed with flexible appendages inspired by a water
  beetle,'' {\em Bioinspiration \& biomimetics}, vol.~14, no.~6, p.~066003,
  2019.

\bibitem{baines2022multi}
R.~Baines, S.~K. Patiballa, J.~Booth, L.~Ramirez, T.~Sipple, A.~Garcia,
  F.~Fish, and R.~Kramer-Bottiglio, ``Multi-environment robotic transitions
  through adaptive morphogenesis,'' {\em Nature}, vol.~610, no.~7931,
  pp.~283--289, 2022.

\bibitem{palmisano2013power}
J.~S. Palmisano, J.~Geder, M.~Pruessner, and R.~Ramamurti, ``Power and thrust
  comparison of bio-mimetic pectoral fins with traditional propeller-based
  thrusters,'' in {\em 18th International Symposium on Unmanned Untethered
  Submersible Technology}, 2013.

\bibitem{diaz2021minimal}
K.~Diaz, T.~L. Robinson, Y.~O. Aydin, E.~Aydin, D.~I. Goldman, and K.~Y. Wan,
  ``A minimal robophysical model of quadriflagellate self-propulsion,'' {\em
  Bioinspiration \& Biomimetics}, vol.~16, no.~6, p.~066001, 2021.

\bibitem{simha2020flapped}
A.~Simha, R.~Gkliva, {\"U}.~Kotta, and M.~Kruusmaa, ``A flapped paddle-fin for
  improving underwater propulsive efficiency of oscillatory actuation,'' {\em
  IEEE Robotics and Automation Letters}, vol.~5, no.~2, pp.~3176--3181, 2020.

\bibitem{wang2024aquamilr}
T.~Wang, N.~Mankame, M.~Fernandez, V.~Kojouharov, and D.~I. Goldman,
  ``Aquamilr: Mechanical intelligence simplifies control of undulatory robots
  in cluttered fluid environments,'' {\em arXiv preprint arXiv:2407.01733},
  2024.

\bibitem{wang2023mechanical}
T.~Wang, C.~Pierce, V.~Kojouharov, B.~Chong, K.~Diaz, H.~Lu, and D.~I. Goldman,
  ``Mechanical intelligence simplifies control in terrestrial limbless
  locomotion,'' {\em Science Robotics}, vol.~8, no.~85, p.~eadi2243, 2023.

\bibitem{chong2023multilegged}
B.~Chong, J.~He, D.~Soto, T.~Wang, D.~Irvine, G.~Blekherman, and D.~I. Goldman,
  ``Multilegged matter transport: A framework for locomotion on noisy
  landscapes,'' {\em Science}, vol.~380, no.~6644, pp.~509--515, 2023.

\bibitem{huang2021cormorant}
J.~Huang, J.~Liang, X.~Yang, H.~Chen, and T.~Wang, ``Cormorant webbed-feet
  support water-surface takeoff: Quantitative analysis via cfd,'' {\em Journal
  of Bionic Engineering}, vol.~18, pp.~1086--1100, 2021.

\bibitem{liu2018platform}
H.~Liu, L.~Shi, S.~Guo, H.~Xing, X.~Hou, and Y.~Liu, ``Platform design for a
  natatores-like amphibious robot,'' in {\em 2018 IEEE International Conference
  on Mechatronics and Automation (ICMA)}, pp.~1627--1632, IEEE, 2018.

\bibitem{pham2020dynamic}
V.~A. Pham, T.~T. Nguyen, B.~R. Lee, and T.~Q. Vo, ``Dynamic analysis of a
  robotic fish propelled by flexible folding pectoral fins,'' {\em Robotica},
  vol.~38, no.~4, p.~699–718, 2020.

\bibitem{sharifzadeh2021reconfigurable}
M.~Sharifzadeh, Y.~Jiang, and D.~M. Aukes, ``Reconfigurable curved beams for
  selectable swimming gaits in an underwater robot,'' {\em IEEE Robotics and
  Automation Letters}, vol.~6, no.~2, pp.~3437--3444, 2021.

\bibitem{sharifzadeh2020curvature}
M.~Sharifzadeh and D.~M. Aukes, ``Curvature-induced buckling for flapping-wing
  vehicles,'' {\em IEEE/ASME Transactions on Mechatronics}, vol.~26, no.~1,
  pp.~503--514, 2020.

\bibitem{johansson2003delta}
L.~C. Johansson and R.~{\AA}. Norberg, ``Delta-wing function of webbed feet
  gives hydrodynamic lift for swimming propulsion in birds,'' {\em Nature},
  vol.~424, no.~6944, pp.~65--68, 2003.

\bibitem{fish1996transition}
F.~E. FISH, ``{Transitions from Drag-based to Lift-based Propulsion in
  Mammalian Swimming1},'' {\em American Zoologist}, vol.~36, pp.~628--641, 08
  2015.

\bibitem{clifton2018comparative}
G.~T. Clifton, J.~A. Carr, and A.~A. Biewener, ``Comparative hindlimb myology
  of foot-propelled swimming birds,'' {\em Journal of anatomy}, vol.~232,
  no.~1, pp.~105--123, 2018.

\bibitem{badri2022birdbot}
A.~Badri-Spr{\"o}witz, A.~Aghamaleki~Sarvestani, M.~Sitti, and M.~A. Daley,
  ``Birdbot achieves energy-efficient gait with minimal control using
  avian-inspired leg clutching,'' {\em Science Robotics}, vol.~7, no.~64,
  p.~eabg4055, 2022.

\bibitem{chatterjee2022multi}
A.~Chatterjee, A.~Mo, B.~Kiss, E.~C. Gonen, and A.~Badri-Spr{\"o}witz,
  ``Multi-segmented adaptive feet for versatile legged locomotion in natural
  terrain,'' {\em arXiv preprint arXiv:2209.08499}, 2022.

\bibitem{chang2017mechanical}
Y.-H. Chang and L.~H. Ting, ``Mechanical evidence that flamingos can support
  their body on one leg with little active muscular force,'' {\em Biology
  Letters}, vol.~13, no.~5, p.~20160948, 2017.

\bibitem{biewener2001dynamics}
A.~A. Biewener and W.~R. Corning, ``Dynamics of mallard (anas platyrynchos)
  gastrocnemius function during swimming versus terrestrial locomotion,'' {\em
  Journal of Experimental Biology}, vol.~204, no.~10, pp.~1745--1756, 2001.

\bibitem{carr2008muscle}
J.~A. Carr, {\em Muscle function during swimming and running in aquatic,
  semi-aquatic and cursorial birds}.
\newblock PhD thesis, Northeastern University, 2008.

\bibitem{grimminger2020open}
F.~Grimminger, A.~Meduri, M.~Khadiv, J.~Viereck, M.~W{\"u}thrich, M.~Naveau,
  V.~Berenz, S.~Heim, F.~Widmaier, T.~Flayols, {\em et~al.}, ``An open
  torque-controlled modular robot architecture for legged locomotion
  research,'' {\em IEEE Robotics and Automation Letters}, vol.~5, no.~2,
  pp.~3650--3657, 2020.

\bibitem{messinger1979mechanics}
D.~S. Messinger, {\em Mechanics of walking and swimming of the duck Anas
  platyrhynchos}.
\newblock PhD thesis, The Ohio State University, 1979.

\bibitem{hepp2022novel}
J.~Hepp and A.~Badri-Spr{\"o}witz, ``A novel spider-inspired rotary-rolling
  diaphragm actuator with linear torque characteristic and high mechanical
  efficiency,'' {\em Soft Robotics}, vol.~9, no.~2, pp.~364--375, 2022.

\bibitem{lin2024bioinspired}
J.~Lin, J.~Ke, R.~Xiao, X.~Jiang, M.~Li, X.~Xiao, and Z.~Guo, ``Bioinspired
  bidirectional stiffening soft actuators enable versatile and robust
  grasping,'' {\em Soft Robotics}, 2024.

\bibitem{taylor2020waddle}
K.~R. Taylor-Burt, {\em How to waddle with a paddle: a study of duck hindlimb
  anatomy, kinematics, and muscle function across behaviors and species}.
\newblock PhD thesis, Harvard University, 2020.

\bibitem{kim2011characteristics}
D.~Kim and M.~Gharib, ``Characteristics of vortex formation and thrust
  performance in drag-based paddling propulsion,'' {\em Journal of Experimental
  Biology}, vol.~214, no.~13, pp.~2283--2291, 2011.

\bibitem{regnault2017}
S.~Regnault, V.~R. Allen, K.~P. Chadwick, and J.~R. Hutchinson, ``Analysis of
  the moment arms and kinematics of ostrich ({Struthio} camelus) double
  patellar sesamoids,'' {\em Journal of Experimental Zoology Part A: Ecological
  and Integrative Physiology}, pp.~163--171, 2017.

\bibitem{prilutsky1994tendon}
B.~I. Prilutsky and V.~M. Zatsiorsky, ``Tendon action of two-joint muscles:
  transfer of mechanical energy between joints during jumping, landing, and
  running,'' {\em Journal of biomechanics}, vol.~27, no.~1, pp.~25--34, 1994.

\bibitem{biewener1989scaling}
A.~A. Biewener, ``Scaling body support in mammals: limb posture and muscle
  mechanics,'' {\em Science}, vol.~245, no.~4913, pp.~45--48, 1989.

\bibitem{ijspeert2007swimming}
A.~J. Ijspeert, A.~Crespi, D.~Ryczko, and J.-M. Cabelguen, ``From swimming to
  walking with a salamander robot driven by a spinal cord model,'' {\em
  science}, vol.~315, no.~5817, pp.~1416--1420, 2007.

\bibitem{sharif2021}
M.~Sharifzadeh, Y.~Jiang, and D.~M. Aukes, ``Reconfigurable curved beams for
  selectable swimming gaits in an underwater robot,'' {\em IEEE Robotics and
  Automation Letters}, vol.~6, no.~2, pp.~3437--3444, 2021.

\bibitem{palmisano2013maximize}
J.~S. Palmisano, R.~Ramamurti, J.~D. Geder, M.~Pruessner, W.~C. Sandberg, and
  B.~Ratna, ``How to maximize pectoral fin efficiency by control of flapping
  frequency and amplitude,'' in {\em 18th International Symposium on Unmanned
  Untethered Submersible Technology}, 2013.

\end{thebibliography}

\end{document}